# "Meet My Sidekick!": Effects of Separate Identities and Control of a Single Robot in HRI

Drake Moore
Northeastern University
Boston, USA
moore.dr@northeastern.edu

Arushi Aggarwal
Northeastern University
Boston, USA
aggarwal.arus@northeastern.edu

Emily Taylor
Northeastern University
Boston, USA
taylor.em@northeastern.edu

Sarah Zhang
Northeastern University
Boston, USA
zhang.sarah1@northeastern.edu

Taskin Padir*
Northeastern University
Boston, USA
Amazon Robotics
Westborough, USA
t.padir@northeastern.edu

Xiang Zhi Tan
Northeastern University
Boston, USA
zhi.tan@northeastern.edu

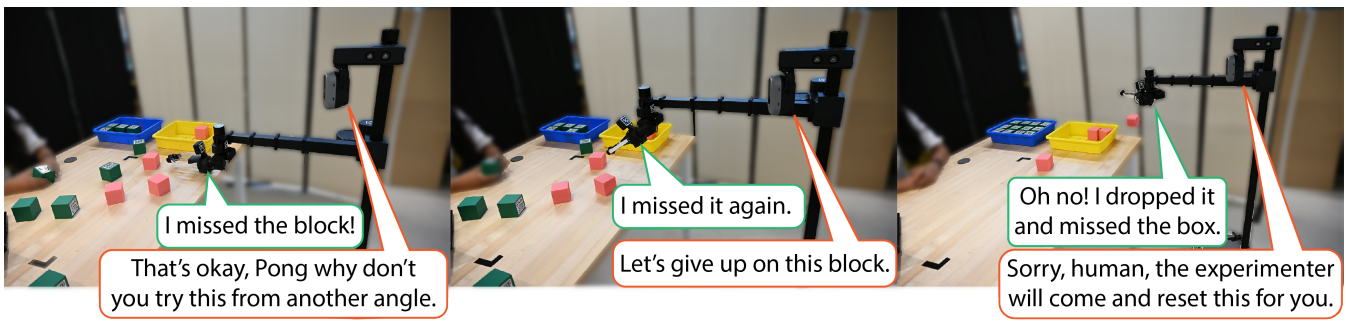

Figure 1: Inter-persona dialogue after errors in the split embodiment condition.

## Abstract

The presentation of a robot's capability and identity directly influences a human collaborator's perception and implicit trust in the robot. Unlike humans, a physical robot can simultaneously present different identities and have them reside and control different parts of the robot. This paper presents a novel study that investigates how users perceive a robot where different robot control domains (head and gripper) are presented as independent robots. We conducted a mixed design study where participants experienced one of three presentations: a single robot, two agents with shared full control (co-embodiment), or two agents with split control across robot control domains (split-embodiment). Participants underwent three distinct tasks – a mundane data entry task where the robot provides motivational support, an individual sorting task with isolated robot failures, and a collaborative arrangement task where the robot causes a failure that directly affects the human participant. Participants perceived the robot as residing in the different control domains and were able to associate robot failure with different identities. This work signals how future robots can leverage different embodiment configurations to obtain the benefit of multiple robots within a single body.

## CCS Concepts

• **Human-centered computing** → **User studies**; **Interaction paradigms**; • **Computer systems organization** → **Robotics**; • **Computing methodologies** → *Multi-agent systems*.

## Keywords

multi-agent human-robot interaction, social robotics, collaborative robotics, embodiment

---

## 1 Introduction

Does one physical robot always equate to interacting with one robot? Aside from the standard single-agent single embodiment,

there have been multiple works in the space of multi-embodiment (see survey [2]) showing how robot identities can move between robot bodies or have multiple identities occupy the same robot body. The fluidity of the presented robot identities can help increase engagement and portray specialization. One multi-embodiment paradigm is co-embodiment, where multiple robot identities reside in the same robot body, and each identity is portrayed as having complete control over the robot's body. While prior work has shown that participants can perceive interacting with multiple identities in different multi-embodiment scenarios, participants still perceive the robot as a single unified physical system. As co-embodiment asks the question of whether there can be multiple identities in a single body, this paper asks whether this concept can be extended – **Can there be multiple robot bodies in a single robot with minimal modification?** Vazquez et al. [30] explored a similar configuration with "Chester" and "Blink" – two physically connected robots with distinct control domains and social cues. Their study found that this setup increased engagement but required custom-built hardware. Our work extends on multi-embodiment research by exploring whether participants can perceive two different robot identities with two different control domains on the same standard robot platform with minimal modifications. We call this setup – **split embodiment**. In our split embodiment setup, we present to the user distinct robot identities that occupy different physical subsystems of a single robot body through the use of identity cues and spatially separated audio output. To the best of our knowledge, this question of distributed control and identity across the robot body has not previously been investigated.

We hypothesize that split embodiment can offer similar benefits to a co-embodiment setup (multiple identities) or multi-robot human interaction setup (multiple physical robots), while additionally enabling compartmentalized failure attribution: failures in one subsystem may not degrade trust in the others.

This could benefit real-world applications like home assistants; for instance, if a robot fails a manipulation task, users might blame the "hand" persona while preserving social engagement with the "head" persona.

This paper presents a novel investigation into split embodiment, specifically examining its affect on user trust and capability assessment during collaborative tasks. We conducted a user study comparing three embodiments: single-agent baseline, co-embodiment (dual agents with shared control), and split-embodiment (separate personas controlling distinct subsystems). We evaluated the effects of these embodiments on tasks involving motivational support, object manipulation, and a sequential arrangement task, additionally exploring how persona-based dynamics during error scenarios influence user perceptions compared to traditional single-agent approaches. We show that participants can not only perceive split-embodiment and correctly associate identities with robot locations, but also appropriately assign blame to the responsible persona.

## 2 Related Works
### 2.1 Multi-Robot Human Interaction
With the increasing affordability and availability of robots, multi-robot human-interaction has become an active area of HRI research, looking into factors such as conformity [21, 23], group perception [7, 8], group dynamics [4–6], and sequential interactions [25, 26].

A multi-robot interaction does not mean all robots must interact with the user. A "passive-social" medium setup is a situation where the user observes the interaction between two robots [22]. Prior work has demonstrated that users are more engaged when observing a dialogue between two robots [9, 10, 24]. Users' perception of the non-interactive robot will also be affected by how it is being treated by the second robot [27].

Nearly all prior work has users interact with multiple distinctive robot bodies. To our knowledge, only one work by Vazquez et al. [30] investigated two distinct robots sharing one body: Chester (a robot wardrobe) with Blink (a lamp) on top, where Blink served as a "funny but mature" sidekick that was found to increase children's engagement. However, Chester and Blink were designed and created specifically to embody a sidekick dynamic. In our work, we achieve this sidekick portrayal on a standard robot platform with minimal modifications (the addition of a $10 USB Bluetooth speaker).

### 2.2 Multi-Robot Identities
Unlike humans, the identities portrayed by a robot system does not have to equal the number of robot bodies. Prior work has explored ways to present robot identity beyond humans' single-identity single-embodiment paradigm. Bejarano et.al. [1], explored how different language presentation cues, such as "self-reference" and "name" can affect users' mental model of a robot team. By having three robots speak at the same time, sharing the same voice, and referring to themselves as "we are", participants perceived all three robots as being controlled by a single identity.

Luria et al. [15] presented four paradigms – one-for-one (a single agent in one robot body), re-embodiment (the shifting of a single agent identity from one body to another [17, 28], also known as agent migration in earlier work [11, 14]), one-for-all (one agent identity controlling multiple robot bodies, also known as hive-mind), and lastly co-embodiment (two or more robot identities controlling the same body) [18]. Our split-embodiment portrays two one-for-one robots controlling different domains of the robot.

Multiple prior works have explored co-embodiment. In [19], the authors examined how agents can follow users around and co-embody a robot system in different scenarios. Reig et.al [18] examined how participants perceive multiple agents that occupy a space habitat and found participants had increased trust when the agents used third-person language to describe systems.

Williams et.al. [31] proposed the Deconstructed Trustee Theory, where robots have both a body and identity "loci of trust" and showed that people could hold different levels of trust for both entities. Our work builds on this work in exploring whether a singular robot body can be split into two distinctly perceived domains that have different levels of trust and perceived capabilities.

## 3 Methods
### 3.1 Study Design
We conducted a mixed-design user study with three embodiment conditions to investigate how identity and control representations affect human perception and collaboration with a single robot system. The study employed a 3x3 mixed design, with embodiment

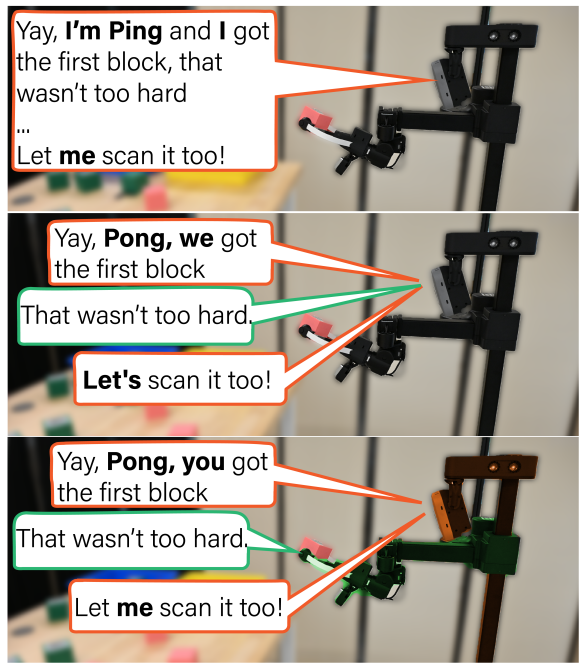

Figure 2: Dialogue was controlled across conditions except for pronoun and name usage to convey intended embodiments. From top to bottom: Single-Agent, Co-Embodiment, and Split-Embodiment

type as a between-subjects factor (single agent baseline, two-agent co-embodiment, two-agent split-embodiment), and collaborative tasks as a within-subjects factor.

## 3.2 Robot Embodiments

We compared three different robot embodiment configurations:

**Single Agent Embodiment:** In the single agent condition, the user perceives a single identifiable agent that controls the robot. This is the common single-robot interaction paradigm and we adopt this approach as our baseline method to compare alternative embodiments against.

**Co-Embodiment:** In the co-embodiment condition, two personas share control of the complete robot system while maintaining distinct identities through different voices and names.

**Split Embodiment:** In the split embodiment condition, the user perceives the controls of a single robot's subsystems as distributed across multiple personas. In this paradigm, two personas with distinct voices and names coexist within a single robot body, with each persona portrayed as maintaining autonomous control over a specific area of the robot. Unlike the standard single-embodiment approaches, split embodiments portray a unique intra-robot collaborative setting to the user, where personas must communicate and negotiate their sub-domain controls while sharing the same body.

Dialogue during the tasks was kept as similar as possible between all three conditions to avoid it being a confound. The baseline condition required a small degree of dialogue restructuring to accommodate single-persona interaction, while co-embodiment differed from split-embodiment only in the usage of the "we" pronoun instead of "I" when self-referencing. The differences across our conditions are described as follows:

(1) **Introduction Protocol:** How the robotic system introduces itself and the tasks.
(2) **Dialogue Pronouns:** The usage of "I" versus "we" when referencing itself/themselves.
(3) **Persona Addressing:** Direct naming when personas address each other.
(4) **Robot Idle Movement:** Minor head/wrist camera movements during speech, to simulate natural human/robot interaction patterns.

## 3.3 Task

The benefit of split embodiment may affect different tasks in different ways. To explore this, we designed three collaborative tasks that examine different aspects of human-robot interaction: (1) a **motivation** task involving minimal physical interaction to establish the robot embodiment without failure scenarios, (2) an independent **sorting** task where robot errors occur but do not affect participant success, and (3) a collaborative **arranging** task where robot failures directly impact participant task completion. In the latter two tasks, planned robot failures were implemented within the manipulation context to examine how participants would attribute errors when different areas of the robot were portrayed as being controlled independently.

*3.3.1 Motivation.* In this task, we investigate whether users would find split-embodiment more engaging/motivating than baseline. Participants were instructed to copy rows of names and quantities into an online form one at a time while the robot simultaneously completed its own "task" while facing away from the participant. Between each interaction, the robot simulated completing its task for a 100 second interval, then rotated to face the participant and started a motivational interaction. In co-embodiment and split-embodiment, the robot dialogue was split across the two personas in a back and forth pattern.

There were four motivation interactions – a motivational speech, a motivational phrase, a joke, and the robot offering an "air low-five" which participants could choose to partake in towards the end of the task. At the end of the task, the experimenter told participants they could stop the task.

This task served as a familiarization period with the robot, allowing participants to observe inter-persona dynamics (when present) and become comfortable with robot behaviors.

*3.3.2 Sorting.* In the sorting task, we investigated fault attribution and perceived capability of the robot in an independent task. Participants and the robot independently sorted blocks laid out on a table into designated stow locations within a bin (Fig. 1). The participants were assigned to organize twelve green blocks, while the robot was assigned four pink blocks. The task involved the user locating a QR code adhered to one side of the block, and scanning it using a laptop's webcam next to the participant. After a 10 second delay, the laptop indicated the associated stow location within the bin for the participant to organize the block into. The robot performed the same task alongside the participant, simulating a scan before organizing the block into its own bin.

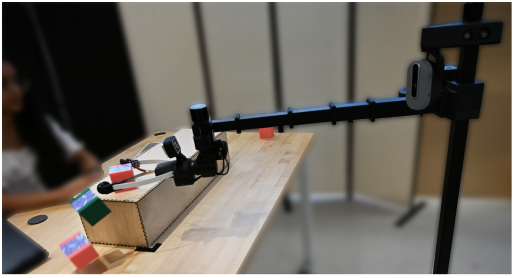

Figure 3: Robot sweeping placed blocks off the table during the Arranging Task

*Planned Robot Failure:* In this task, the robot successfully picked, scanned, and placed the first and third blocks that it was assigned. For the second block, the robot failed to grasp the block twice, before subsequently giving up and moving on (Fig. 1). For the fourth assigned block, the robot successfully grasped the block, but missed the deposit location, dropping it off the side of the table. Following the failure, the grasping robot had either self-assuring statements (baseline) or assurance from the main agent (co-embodiment & split-embodiment).

*3.3.3 Arranging.* In the arranging task, we investigate fault attribution and change in perceived capability of the robot in a collaborative task where the robot's failure directly affects the participants. Participants worked directly with the robot to create a sequential block pattern based on images adhered to each block. The human and the robot were each given three blocks in an alternating order to arrange together. The robot began by laying down the first block, and then indicated to the participant to place the next block in the pattern alongside it. The robot placed three pink blocks on the raised table one at a time, waiting for the participant to place theirs between each block.

*Planned Robot Failure:* After placing the second block, the robot attempted to slide blocks together but instead swept all blocks off the table, disrupting the established pattern (Fig. 3). The robot then requested participant assistance replacing all the knocked blocks back onto the table in the original order, and continued collaboration. During the failure, the robot had either self-assuring statements (baseline) or assurance from the main agent (co-embodiment & split-embodiment).

### 3.4 System Implementation

We use the mobile manipulator Hello Robot Stretch 3 [20] for this study. The platform consists of a two-wheeled differential drive mobile base, a lift, a telescoping arm, and a 2DOF yaw-pitch RealSense D435 camera mounted at the top of the lift. The end effector is a 3DOF yaw-pitch-roll wrist with a compliant gripper and a RealSense D405 camera mounted above it.

Control of the robotic system was implemented using the Robotic Operating System 2 (ROS2) framework [16] for distributed control and communication between system components. Robot manipulation was accomplished with a visual servoing control loop using a fine-tuned You Only Look Once (YOLO) object detection model trained to identify the blocks used in the tasks. Upon detection and depth localization, the system employed inverse kinematics (IK) solving to compute joint trajectories for object manipulation.

All robot dialogue was generated using Coqui's VITS (Conditional Variational Autoencoder with Adversarial Learning for End-to-End Text-to-Speech) text-to-speech model [12], providing natural speech generation for verbal interactions with participants. In conditions with multiple embodiments, the second agent's speech was sampled at a slightly higher rate, eliciting a distinguishable higher-pitched voice. We chose this over unique voices to prevent voice type impacting participants' perception of a persona.

To avoid the need for complex action recognition models, any robot actions that required human users to complete an action (e.g. air low-five, participant placing blocks during arranging) were secretly triggered by an out-of-sight experimenter monitoring the task to begin the robot's next action when appropriate.

**Implementation of co-embodiment:** To represent the co-embodiment condition, both agents spoke through the robot's head-mounted speaker and performed idle movement using its "head", communicating shared system control to the participant.

**Implementation of split-embodiment:** To represent the split-embodiment condition, we opted for one persona to demonstrate control of the "head" (2DOF yaw-pitch camera) and robot base, speaking through the head-mounted speaker with associated head idle movements. The second persona displayed control of the lift, telescopic arm, and 3DOF end-effector (shown in Fig. 2), and spoke through a wrist-mounted speaker with associated idle movements in the wrist. This division created two semi-autonomous agents controlling two spatially distinct subdomains, while maintaining a coherent representation of the complete robotic system.

**Robot dialogue:** In our study, all dialogue was pre-scripted and pre-generated to match robot actions. For the co-embodiment and split-embodiment conditions, the first persona was named "Ping" and the second was "Pong". We chose these names to avoid automatic role assignment (names like "Heady" or "Handy" would allude to specific body regions), and to prevent bias if one name is easier to remember. To ensure consistency between conditions, "Pong" was the one that talked in reference to all manipulation actions. The dialogue was kept similar between the conditions, with the differences being pronoun usage, self-referencing, and idle motions [1].

### 3.5 Hypotheses

Based on prior work on multi-identity trust [31] and multi-robot human interactions [22], we have the following five hypotheses on the effects of split-embodiment.

H1 - Participants will be able to distinguish the two agents as having separate control over different control domains in the split-embodiment condition.

H2 - **Motivation** (a) Both multi-agent conditions will be perceived as more warm than the baseline and reduce boredom during the task. (b) Split-embodiment will see further improvements in the warmth and boredom metrics compared to co-embodiment.

---
[1] the complete dialogue between agents are included in the supplementary materials

- H3 - **Sorting** Participants will perceive Ping in the split-embodiment as more competent than in the co-embodiment and baseline conditions. Pong's competence and trustworthiness will decrease while Ping's stays the same.
- H4 - **Arranging** Participants will perceive Ping as more competent and trustworthy in the split-embodiment than in the co-embodiment and baseline conditions. Pong's competence and trustworthiness will decrease while Ping's stays the same.
- H5 - Participants will assign blame to Ping in the baseline condition, both personas in the co-embodiment condition, and to only Pong in the split-embodiment condition.

### 3.6 Participants

45 participants were recruited through fliers posted on our university campus. 9 participants were excluded due to unplanned robotic system failure or user errors when completing the post-task surveys. This left us with 36 participants deemed usable for further analysis (12 participants in each embodiment condition). Included participants had an average age of 22.69 (SD = 4.02). On a scale of 1 (lowest) to 7 (highest), their mean scores for experience with computers and robots were 7 (SD = 0) and 5 (SD = 1.66), respectively. The study took about 60 minutes and participants were given a 15 USD Amazon gift card for participating. This study was approved by Northeastern University's Institutional Review Board.

### 3.7 Procedure

After explaining the study tasks and obtaining consent, the experimenter explained to the participant that the robot would introduce itself/themselves as well as the task, and then indicate to the participant when to begin the task. After the persona(s) introduced each task, the experimenter answered any of the participant's questions regarding the task before beginning the experimental scenario.

All participants started with the motivation task to avoid it being influenced by robot failure. After the motivation task, the participants completed arranging and sorting task. The order of arrange and sort was counterbalanced across participants.

### 3.8 Measures

Participants completed surveys between each task to assess their overall perception of the robot. Social perception was measured using a Robotic Social Attributes Scale (RoSAS) questionnaire [3] on a 9-point Likert scale (1 = strongly disagree, 9 = strongly agree), while robot trust was evaluated using a Multi-Dimensional Measure of Trust (MDMT) questionnaire [29] on a 7-point Likert scale (1 = strongly disagree, 7 = strongly agree) with a not applicable option.

Additionally, participants answered their levels of wariness about each persona and their potential to be injurious or cause harm following each task [2]. We aimed to see whether progressive mistakes would increase wariness about the robot persona(s) overall. Following the sorting and arranging tasks, in which mistakes were intentionally made by the robot, participants were asked to answer which persona bore the responsibility for any mistakes that occurred. They could choose between either of the persona(s), or that both or neither was to blame.

---

[2] All custom questions are included in the supplementary materials

During the co-embodiment and split-embodiment conditions, participants provided additional assessments on their perceived relationship between the two agents via six custom questions represented on a 7-point Likert scale. There was also an additional open-ended question that allowed participants to express their impression of the dynamic.

Following all three tasks, participants in all three conditions were asked to highlight the regions of the robot that they thought were occupied by each of the persona(s). The aim was to determine whether participants could distinguish between personas and their control over the robot's subsystems.

Finally, demographic information was collected, including self-reported computer usage and robot familiarity.

## 4 Results

We analyzed participant survey responses using Bayesian repeated measures ANOVAs to examine how embodiment affects human-robot interaction measures. Because Pong does not exist in the single-agent condition, we have two analysis models for repeated survey measures. The first model included two within-subjects factors (Task and Persona) and one between-subjects factor (Embodiment Condition – co-embodiment and split-embodiment only). The second model keeps the Persona to only Ping but has one within-subjects factor (Task) and one between-subject factor (the three embodiment conditions). In all cases, we only report the best model with the highest Bayes factor if it's at least anecdotal ($BF_{10} > 3.33$). Except for the RoSAS Discomfort subscale for Pong in the sorting task ($\alpha = 0.67$), all our subscales have at least an acceptable level of reliability [13] with a Cronbach's alpha score of >0.74.

### 4.1 Persona Identification

When asked where the identities reside (Fig. 4), participants predominantly assigned Ping to the head, and Pong to the gripper in both the split-embodiment and co-embodiment conditions (n=9, 75% and n=7, 58.3% respectively). We conducted Fisher's exact tests which confirmed significant differences across the complete distribution of responses (p=0.0004). Pairwise comparisons revealed that participants in the split-embodiment condition consistently distinguished head from gripper attributions (p=0.0003), while those in the co-embodiment condition did not (p=0.20). 1 participant (8.3%) in split-embodiment and 3 participants (25%) in co-embodiment identified Ping as inhabiting the gripper and Pong as inhabiting the head. A small subset in the co-embodiment condition identified both personas as inhabiting both the head and the gripper (n=2, 16.7%). Interestingly, two participants in the split-embodiment condition did not identify Ping as residing in the head or the gripper (n=2, 16.7%), with one additionally indicating that Pong did not reside in the head or gripper as well (n=1, 8.3%).

### 4.2 Robot Warmth, Discomfort, and Interaction

There was no to little evidence ( $0.33 < BF_{10} < 3$ ) that the tasks, personas, or embodiment conditions affected the warmth and discomfort RoSAS subscales.

For the task measures, we found no evidence for or against the embodiment condition having an effect on boredom ($BF_{10} = 0.39$) during the motivation tasks. We found anecdotal evidence ($BF_{10} =$

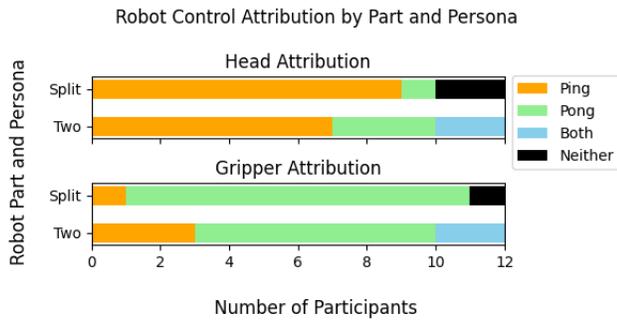

Figure 4: Participant attributions of who controls the head and gripper across co-embodiment and split-control conditions. As indicated in the graph, one participant in the split condition and three in the co-embodiment condition swapped Ping and Pong's personas when attributing control.

0.24) against the embodiment condition having an effect on the enjoyment of the motivation task.

We found no evidence for or against ($BF_{10} = 0.38$) the embodiment condition having an effect on the perceived relationship between Ping and Pong.

### 4.3 Robot Capability Assessment

*MDMT Reliability subscale:* We found moderate evidence that experimental task affected participants' assessment of robot reliability ($BF_{10} = 4.85$). Post-hoc analysis revealed strong evidence for decreased reliability from motivate (M=4.50, SD=1.57) to sort (M=3.69, SD=1.61), $BF_{10} = 18.78$, and moderate evidence for a decline from moderate to arrange (M=3.89, SD=1.46), $BF_{10} = 7.53$. However, anecdotal evidence suggested no difference between sort and arrange ($BF_{10} = 0.25$). All other factors were inconclusive ($BF_{10} < 3.33$).

*MDMT Competence subscale:* In our main model, we found strong evidence ($BF_{10} = 14.14$) that the type of task affected the agent's perceived competence. Post-hoc analysis showed very strong evidence that agent competence rating in sort (M=3.782, SD=1.866) and arrange (M=4.05, SD=1.63) was lower than motivate (M=4.822, SD=1.182). ($BF_{10} = 38.74$ and $BF_{10} = 54.30$). There is anecdotal evidence against a difference between sort and arrange ($BF_{10} = 0.32$). All other factors were inconclusive ($BF_{10} < 3.33$).

*RoSAS Competence Subscale:* Our main model revealed moderate evidence that task influence participants' perceptions of robot competence ($BF_{10} = 3.61$). Post-hoc comparisons demonstrated strong evidence for competence decline from motivate (M=5.92, SD=1.78) to sort (M=5.10, SD=1.55), $BF_{10} = 15.53$, with moderate evidence for decline from motivate to arrange (M=5.32, SD=1.74), $BF_{10} = 5.21$. Anecdotal evidence indicated no meaningful difference between sort and arrange ($BF_{10} = 0.27$). All remaining factors yielded inconclusive evidence ($BF_{10} < 3.33$).

*Single Embodiment Condition Findings:* Surprisingly, the single agent embodiment showed no evidence of a decrease in competence or trust ratings across the experimental tasks ($BF_{10} < 3.33$), despite experiencing identical planned failure scenarios as the multi-agent embodiment conditions. This finding suggests that the attribution and impact of robot failures may be directly altered by the presence of multiple distinct personas within the system. Future works should investigate this effect further.

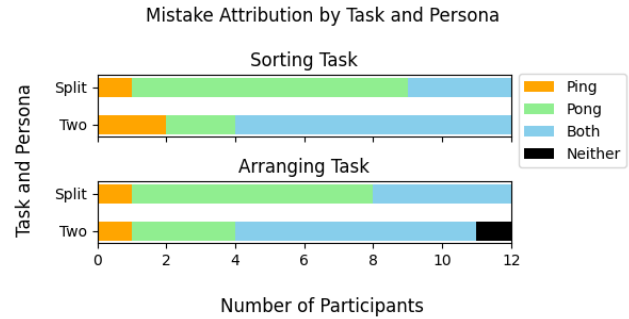

Figure 5: Participant attributions of who was to blame for the mistake made in each task across co-embodiment and split-control conditions.

### 4.4 Mistake Attribution

In the sorting task, the attribution of mistakes for task failures varied significantly across experimental conditions according to a Fisher's Exact Test ($p = 0.04$). Participants in the split-embodiment condition largely assigned the mistake to Pong (n=8, 67%), compared to the co-embodiment condition where participants predominantly attributed the mistake to both agents (n = 8, 67%).(Fig. 5).

However, the attribution of mistakes did not differ significantly in the Arranging Task ($p = 0.29$). In the split-embodiment condition, the majority (n=7, 58%) of the participants attributed the mistake to Pong, with the remaining attributing it to both agents ($n = 4$) or Ping ($n = 1$). In the co-embodiment condition, the majority (n = 7, 58%) of the participants attribute the mistake to both agents. The remaining participants attributed it to either Ping ($n = 1$), Pong ($n = 3$), or Neither ($n = 1$)

In the single-agent condition, all participants ($n = 12$) attributed the mistakes to Ping in both the sorting and arranging tasks.

### 4.5 User Responses

Participants in the co-embodiment and split-embodiment conditions were asked to describe their perception of the relationship between Ping and Pong at the end of the experiment. Two researchers then labeled these responses into Hierarchy, Together, and Other categories (91.6% agreement). Differences were resolved through discussion, opting for the stricter exclusion from Hierarchy and Together between the two. There were no significant differences found between the multi-agent embodiment conditions. Interestingly, even though the split embodiment condition introduced Pong as Ping's "sidekick", more participants perceived them to be collaborating equally rather than having implemented hierarchy.

## 5 Hypothesis Support

H1 **Identification of Split-Embodiment - [Supported]** We found evidence that supported Hypothesis 1. Participants demonstrated a significant difference in the ability to differentiate between personas based on their association with the two distinct robot control domains (arm and head). This means that participants can perceive two identities in different parts of the robot by simply adding a speaker, and language and voice cues.

|                  | Hierarchy | Together | Other |
|------------------|-----------|----------|-------|
| Co-embodiment    | 5         | 5        | 2     |
| Split-embodiment | 3         | 5        | 4     |

Table 1: Count of labeled participant relationship of Ping and Pong response categories.

H2 **Effects on Motivation Task - [Not Supported]** We found no evidence that the embodiment condition had an effect on the perceived warmth for either identity. While there was anecdotal evidence that the embodiment condition does not affect task enjoyment, we believe it is too weak to draw any conclusions.

H3 **Competence & Trust in Sorting - [Not Supported]** We found no evidence that the embodiment condition affects the robot's perceived competence and trust in the sorting task.

H4 **Competence & Trust in Arranging - [Not Supported]** We found no evidence that the embodiment condition affects the robot's perceived competence and trust in the arranging task.

H5 **Mistake attribution - [Partially Supported]** We found that mistake attribution was significantly different only in the sorting task, but not in the arranging task. In the sorting task, participants in the split-embodiment condition attributed the mistake mostly to Pong, who was in control of the hand, while participants in the co-embodiment condition attributed the mistake to both robots.

While the majority of participants in the arranging task also attributed mistakes to Pong in split-embodiment and both in co-embodiment, this trend was not significant. We believe this was due to the way the failure in arranging the task was designed: the sweeping of the blocks required both gripper movement and the robot base spinning. This minor difference may have made participants perceive the mistake as a combined error. This variation in mistake attribution provides some evidence that separate embodiments affect not only how participants perceive robot competence, but also how they assign responsibility following robot errors. The concentrated attribution of robot failures to Pong in the split-embodiment condition compared to the joint attribution in co-embodiment reveals how physical embodiment shapes responsibility assignment. Split-embodiment better enables compartmentalization of robot failures to specific subdomains based on participant's mental models, while co-embodiment's shared physical representation distributes blame across both agents.

## 6 Exploratory Analysis

We conducted additional exploratory analyses to explore two study limitations. This should be interpreted as preliminary evidence requiring further validation with improved methodology.

### 6.1 Temporally Aligned Analysis

Our original analysis did not investigate the compound effects of both failures. We conducted an additional exploratory analysis on the difference between the motivation task (before failures) and the completion of the study (after failures). We temporally aligned participant responses according to their task ordering, labeling the first task as "Motivate" and the final collaborative task as "Task 2". Following this alignment, we conducted the same analysis described in the Results section. Analyses comparing post-motivate and post-task 2 were in line with the findings from the Results analysis, but with a higher Bayes factor (e.g., Competence from $BF_{10}$ = 14.14 to $BF_{10}$ = 17.04). This is likely due to participants observing multiple robot failures by the end of the experiment.

### 6.2 Embodiment Aligned Participant Responses

We also suspect a small number of participants (3 in co-embodiment, 1 in split-embodiment) mixed up the identities and inverted the embodiment mapping, labeling Ping as controlling the wrist and Pong as controlling the head, which is the reverse of our design and what other participants responded to. This was further supported by mistake attribution and qualitative responses (P44:"... Pong to be a reviewer and Ping acting according to that"). To investigate embodiment effects on these personas, we swapped these four participants' Ping and Pong responses to match the majority and repeated the temporally aligned analysis described above.

*MDMT Reliability Subscale:* The model revealed strong evidence that the model with the strongest support included Task, Persona, and their interaction ($BF_{10}$ = 12.93). This suggests the task might have influenced participants' perception of robot reliability based on the persona (Ping or Pong). Post-hoc Bayesian paired sample t-tests were conducted to explore this interaction. An extreme difference in performance was observed between motivate (M=4.53, SD=1.55) and task 2 (M=3.10, SD=1.53) under Pong ($BF_{10}$ = 182.33), while no significant difference was found under Ping ($BF_{10}$ = 0.55).

*RoSAS Competence Subscale:* The model comparison revealed that the model including the main effects of Task and Persona, as well as their interaction (Task × Persona), received extreme support from the data relative to the null and alternative models ($BF_{10}$ = 100.13). This indicates a strong interaction between task and personas affecting user's assessment of the persona's capabilities. Post-hoc Bayesian paired sample t-tests revealed that under Pong, there was very strong evidence ($BF_{10}$ = 35.39) for a difference between motivate (M=5.91, SD=1.77) and task 2 (M=4.62, SD=1.54). Under Ping, the evidence for a task effect was anecdotal ($BF_{10}$ = 0.46), indicating that the interaction was primarily driven by differences in responses to Pong.

Future studies should employ more distinctive names and explicit manipulation checks for name-region alignment to better examine split-embodiment effects on trust and competence perceptions.

## 7 Discussion

### 7.1 Perception of Multi-agent Embodiment in Robot Systems

Our results demonstrate that participants could successfully differentiate personas across all embodiment conditions, despite the differences being limited to a brief introduction, minor dialogue, and speech idle movements, and a slight pitch difference between voices.

Even without distinct embodiment cues in the co-embodiment condition, participants attributed separate identities to different robot control domains based solely on language and action cues.

This suggests that, in addition to identity, robot control domain might be an additional parameter to understand the interplay between identity and body. Suppose when a robot's identity in either co-embodiment or re-embodiment only uses part of the robot, people may perceive that robot identity as only living in that particular part of the body.

### 7.2 Embodiment Effects on Failure Perception

Our findings reveal two interconnected effects of embodiment configuration on how participants process robot failures: the distribution of responsibility attribution and the asymmetric impact on persona-specific capability and trust assessments.

**Responsibility Attribution:** Embodiment configuration directly shaped how participants assigned responsibility following robot failures. Our results provide evidence that split-embodiment better enables participants to compartmentalize robot failures to specific robot identities and domains, while co-embodiment's shared physical representation distributes blame across both agents. The unanimous attribution of mistakes to Ping in the single agent condition provides an important reference point, demonstrating that the presence of multiple personas, regardless of embodiment configuration, fundamentally changes responsibility attribution patterns. While we demonstrate that blame assignment happens to the persona, more work is needed to explore whether this attribution is primarily to the body or the identity [31].

**Asymmetric Capability and Trust Erosion:** Even though we found no support for H3 and H4, our embodiment-aligned exploratory analysis showed some preliminary evidence that there is asymmetry in how task failures affected the two personas. Pong, who controlled the manipulator and was directly responsible for the planned failures in the split-embodiment condition, experienced declines in perceived competence and reliability trust. In contrast, Ping's ratings showed only anecdotal to moderate evidence of decline. This shows preliminary evidence that participants constructed different mental models of each persona's competence rather than assessing the robot system as a whole.

### 7.3 Design Implications for Split-Embodiment

**Implication 1:** Our findings suggest that split-embodiment may provide a form of blame compartmentalization, where failures in one control domain do not necessarily compromise trust in other domains within the same robot system. This paradigm has the potential to enable error-stable human-robot partnerships, especially in complex tasks requiring diverse capabilities where localized robot failures are inevitable. This compartmentalization effect can be leveraged in robot system designs by deliberately assigning failure-prone capabilities to a smaller, specialized persona, allowing the overall system to retain user trust.

**Implication 2:** Our findings also show that designers of multi-embodiment systems should be aware of the control domain. If an identity always controls a particular part of the robot, participants may perceive that the robot only resides in that part of the body and focus any social or trust perception to that area.

### 7.4 Limitations and Future Work

Several limitations constrain the generalizability of our findings. As a novel work investigating various embodiment effects in direct HRI, our study employed constrained dialogue scripts to isolate embodiment effects and other confounds, limiting the experiment from examining how various inter-persona dynamics and interaction may influence perceptions. In two sessions, participants tried talking to our robot and got no responses. Future work should investigate how inter-persona dynamics, disagreements, and explicit coordination between personas affect overall embodiment effects on user perception. We also draw from a college-age student crowd with high familiarity with computers and robots. Replication of our study should be conducted with the broader population or children when they are developing theory of mind.

The phonetically similar names "Ping" and "Pong" caused confusion—several participants requested clarification during post-task surveys, reporting difficulty remembering which voice corresponded to which name. This led to our attempt to correct the confusion in the exploratory analysis. Future work should use more phonetically distinct names to prevent this issue.

Our study employed a single robot platform, the Hello Robot Stretch 3, with a particular design architecture that may have facilitated the separation between head-based and arm-based personas in ways that may not generalize to other robots. Future work should examine whether these embodiment effects can be replicated across various robot platforms.

Lastly, the planned failures in our study were relatively benign, potentially reducing the impact on robot trust compared to higher-stakes failures. Understanding how blame compartmentalization functions under more significant failure conditions must be investigated before deploying this strategy in real-world environments. Additionally, our study did not investigate long-term impacts that robot embodiments may have on human-robot relationships. Longitudinal studies examining how mental models of multi-persona systems evolve over time can provide valuable insights when designing robot systems intended for real-world deployment.

## 8 Conclusion

This work demonstrated split-embodiment and proposed control domains as an additional design parameter in multi-robot human interaction. Our study challenges fundamental assumptions about robot identity and inhabitance, and directly impacts human perceptions during human-robot interaction. Our findings demonstrate that people naturally construct unique mental models of robot capability when presented with multiple personas, and that embodiment configuration shapes how they attribute responsibility when those capabilities fail. This asymmetric impact of failures on persona perceptions opens new design paradigms for collaborative robotics for real-world environments.

## 9 Acknowledgement

We thank Samantha Reig, Risha Ranjan, members of the PARCS lab, and the study participants for making this work possible.